\PassOptionsToPackage{table,dvipsnames}{xcolor} 

\documentclass{article} 
\usepackage{collas2025_conference,times}
\usepackage{easyReview}

\usepackage{amsmath,amsfonts,bm}

\def\eqref#1{equation~\ref{#1}}

\def\1{\bm{1}}

\DeclareMathAlphabet{\mathsfit}{\encodingdefault}{\sfdefault}{m}{sl}
\SetMathAlphabet{\mathsfit}{bold}{\encodingdefault}{\sfdefault}{bx}{n}

\usepackage{hyperref}
\hypersetup{
    colorlinks=true,
    linkcolor=red,
    filecolor=magenta,
    urlcolor=blue,
    citecolor=purple,
    pdftitle={Self-Regulated Neurogenesis for Online Data-Incremental Learning},
    pdfpagemode=FullScreen,
    }

\title{Self-Regulated Neurogenesis for \\ Online Data-Incremental Learning}

\author{%
  Murat Onur Yildirim$^1$, 
  Elif Ceren Gok Yildirim$^1$, 
  Decebal Constantin Mocanu$^2$, 
  Joaquin Vanschoren$^1$ \\
  $^1$Eindhoven University of Technology, $^2$University of Luxembourg\\
}

\collasfinalcopy 

\usepackage{enumitem}
\usepackage{algorithm}
\usepackage{algorithmic}
\usepackage{multirow}
\usepackage{graphicx}
\usepackage{subcaption}
\usepackage{wrapfig}
\usepackage{xcolor}

\begin{document}

\maketitle

\begin{abstract}
Neural networks often struggle with catastrophic forgetting when learning sequences of tasks or data streams, unlike humans who can continuously learn and consolidate new concepts even in the absence of explicit cues. 
Online data-incremental learning seeks to emulate this capability by processing each sample only once, without having access to task or stream cues at any point in time since this is more realistic compared to offline setups, where all data from novel class(es) is assumed to be readily available.
However, existing methods typically rely on storing the subsets of data in memory or expanding the initial model architecture, resulting in significant computational overhead.
Drawing inspiration from ‘self-regulated neurogenesis’—brain's mechanism for creating specialized regions or circuits for distinct functions—we propose a novel approach SERENA which encodes each concept in a specialized network path called ‘concept cell’, integrated into a single over-parameterized network. Once a concept is learned, its corresponding concept cell is frozen, effectively preventing the forgetting of previously acquired information.
Furthermore, we introduce two new continual learning scenarios that more closely reflect real-world conditions, characterized by gradually changing sample sizes. Experimental results show that our method not only establishes new state-of-the-art results across ten benchmarks but also remarkably surpasses offline supervised batch learning performance. 
The code is available at \url{https://github.com/muratonuryildirim/serena}.
\end{abstract}

\vspace{10pt}
\section{Introduction}
Artificial neural networks have achieved great success, often surpassing human performance in various applications, particularly when operating on datasets assumed to be sampled from independent and identically distributed (iid) static sources. However, real-world systems face a different reality, where they are exposed to sequential streams of non-stationary (non-iid) data. As the input data changes, previously learned weights are overwritten, leading to \textit{catastrophic forgetting}~\citep{catastrophic} which poses a major obstacle to effective continual learning (CL) of sequential streams.

In contrast, humans excel at CL by seamlessly assimilating and integrating novel information from a dynamic and continuous stream of diverse experiences throughout life's continuum. This ability is underpinned by a sophisticated set of neurophysiological processing principles, also called \textit{plasticity-stability} mechanisms~\citep{neuro_cl1, neuro_cl2}.
By continuously monitoring sensory inputs and internal representations, the human brain detects deviations from expected patterns or environmental contexts, signaling concept drift~\citep{neuro_drift0, neuro_drift1}. In response, it triggers adaptive mechanisms such as updating specialized regions known as \textit{concept cells}~\citep{conceptcell}, reallocating attentional resources, or initiating new learning processes to accommodate an evolving environment~\citep{brain_adjusting, brain_drift_detection}. Collectively, these processes contribute to \textit{self-regulated neurogenesis}~\citep{neurogenesis, neurogenesis1}, enabling continuous adaptation and knowledge refinement.

Inspired by these intricate mechanisms, we introduce SERENA, a model that learns each concept within a specialized neural circuit (or network path), integrated into a single over-parameterized network. It does not require multiple epochs of training and explicit task cues by automatically detecting concept changes in the data stream thereby can operate in online data-incremental learning scenarios~\citep{cope}---also referred to as task-free~\citep{task_free} or task-agnostic continual learning~\citep{task_agnostic}. Once a concept is learned, its corresponding concept cell is frozen to prevent forgetting. During inference, rather than simply relying on a single neural path, SERENA adopts another neuro-inspired strategy that incorporates a built-in recency effect~\citep{recency_effect} with an ensemble approach. This prioritizes the recent experiences in the decision-making process while still integrating past knowledge, enhancing adaptability to evolving data.

\newpage
\begin{figure}[h]
    \centering
    \includegraphics[width=0.98\textwidth]{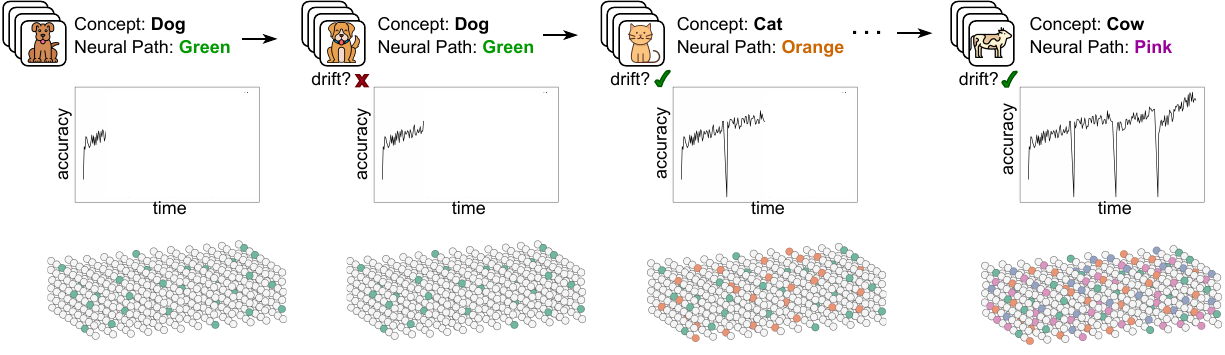} 
    \caption{SERENA assigns random network paths or concept cells~\citep{conceptcell} integrated into a single over-parameterized network at each concept drift that is detected automatically without explicit indicators, mirroring the self-regulated neurogenesis~\citep{neurogenesis, neurogenesis1} without any model growth.}
    \label{fig:teaser}
\end{figure}

Our contributions can be summarized as follows:
\vspace{-5pt}

\begin{enumerate}[label=\Roman*.]
    \item \textbf{SERENA:}
    We introduce a novel online data-incremental learning approach inspired by the brain’s mechanisms, leveraging concept cells formed through zero-cost random pruning during training and a recency-biased ensemble strategy during inference. It offers a simple yet effective solution by continually fine-tuning specific neural paths within a fixed backbone without relying on experience replay or network expansion.

    \item \textbf{New SOTA:} 
    We conduct extensive experiments across eleven benchmarks, demonstrating that SERENA significantly outperforms existing state-of-the-art methods while reducing complexity. Notably, it improves the accuracy of state-of-the-art methods more than $2\times$ on Split-CIFAR100 while even surpassing the performance of traditional iid offline supervised batch learning.
    
    \item \textbf{New CL Scenarios:} 
    We also propose two new continual learning scenarios that more closely reflect real-world dynamics with gradually increasing or decreasing sample sizes throughout learning sessions, complementing the long-tailed CL scenario~\citep{long_tail} and offering a challenging testbed for evaluating continual learning algorithms.
\end{enumerate}
\section{Related Work}

\paragraph{Continual Learning.}
Continual learning approaches primarily target task-incremental learning (TIL) or class-incremental learning (CIL) settings, where models are typically aware of task identifiers during training, testing, or both. These approaches often assume that models can be trained over multiple epochs with repeated shuffling. Regularization-based techniques focus on consolidation~\citep{ewc, si, r_walk, mas, lee2020continual} or knowledge distillation~\citep{lwf, hou2018kd, kang2022kd} to maintain model stability. Replay-based methods either store~\citep{icarl, gem, bic, wa, cls-er, esmer} or generate~\citep{ decebal2016gen, shin2017continual, he2018exemplar, hu2019overcoming, fetril} samples to preserve past experiences. Architecture expansion methods~\citep{expertgate, aanets, dualnet, der, foster, par, dytox, zhou2022model} add completely new layers or networks to enhance model plasticity and parameter isolation techniques use mask learning~\citep{piggyback}, iterative pruning~\citep{packnet, clnp, supsup, cps}, or dynamic sparse training~\citep{spacenet, nispa, wsn, sparcl, afaf, softsubnet, osn} to retain critical parameters. 
Recent research has integrated parameter-efficient fine-tuning (PEFT) techniques into continual learning to reduce task interference by training only small sets of task-specific parameters on top of the large pre-trained backbone~\citep{ease, l2p, codaprompt, tosca}.
Despite their success, these methods often face practical limitations such as the need for explicit task identifiers, the computational burden of multiple training epochs or pre-trained architectures, which underscores more efficient strategies.

\paragraph{Online Data-Incremental Learning.} 
A more realistic direction argues that realistic continual learners should be capable of processing novel data streams in the absence of explicit task cues at any time point.
This concept was first explored in~\citep{task_free} and later extended by MIR~\citep{mir}, which enhanced reservoir sampling with a loss-based retrieval strategy.
Reservoir~\citep{reservoir} serves as a replay baseline with a strong potential to surpass continual learning methods~\citep{reservoir_good_cl_baseline}.
GSS~\citep{gss} extends the optimization perspective of GEM~\cite{gem}, which necessitates prior knowledge of the number of tasks and task transitions, to an instance-based level and adds samples to the buffer based on their gradients.
CoPE~\citep{cope} synchronizes the latent space with continually evolving prototypes with a high momentum-based update to facilitate learning.
CN-DPM~\citep{cndpm} uses a Dirichlet process-based expansion mechanism aiming to increase the model’s capacity, and SEDEM~\citep{sedem} improves the expansion process by introducing a self-assessment mechanism that considers the knowledge diversity among existing modules. Online-LoRA~\citep{onlinelora} uses loss dynamics to detect distribution shifts and fine-tunes pre-trained Vision Transformer models using a novel online weight regularization strategy to prevent forgetting. DCM~\citep{dcm} presents a novel memory management approach that dynamically adjusts memory clusters based on knowledge discrepancy criteria without task-specific signals. However, the practice of storing a subset of data for replay, expanding the network, or relying on large pre-trained architectures presents scalability challenges and is often constrained by factors such as privacy, memory limitations, and computational resources.

\paragraph{Network Pruning.}
Pruning removes some connections of the network and thereby limits the training process to a subset of parameters or neural paths.~\cite{mocanu_topological} showed that pruning a restricted Boltzmann machine at initialization achieves comparable performance to its dense counterpart. Similarly,~\cite{randompruning} demonstrated the effectiveness of randomly pruned backbones when they are sufficiently wide, while~\cite{why_random_all_need} proved that randomly pruned random networks possess ample expressive capacity.
Uniform~\citep{uniform_init1, uniform_init2, uniform_init3}, Erd\H{o}s-R\'{e}nyi (ER)~\citep{set}, and Erd\H{o}s-R\'{e}nyi Kernel (ERK)~\citep{rigl} are primary practical studies that rely on predetermined sparsity ratios but work well in practice.  Recent research proposes more advanced approaches to adjust layer-wise sparsity ratios before training rather than relying on predetermined sparsity ratios~\citep{snip, grasp, synflow}.
Instead of relying on weight magnitudes, edge-popup~\citep{edge_popup} assigns distinct scores to each weight to guide the pruning process. This pruning strategy is employed by an offline continual learning approach SupSup~\citep{supsup} to construct subnetworks. However, it necessitates multiple forward and backward passes to determine the appropriate subnetwork for each task, limiting its applicability to online settings that demand each sample to be processed only once.

\section{Self-Regulated Neurogenesis (SERENA)}
\vspace{-5pt}
In this section, we first share the background of existing online data-incremental learning methods and then provide our research question that motivates us to propose SERENA. Following, we give preliminaries and the details of training and inference of our approach which is illustrated in Figure \ref{fig:teaser} and positioned its novelty in Table \ref{tab:methods}.

\begin{wraptable}[11]{l}{0.45\textwidth}
\vspace{-13pt}
\captionsetup{font=small}
\caption{Online data-incremental learning baselines based on key characteristics.}
\fontsize{8}{10}\selectfont
\label{tab:methods}
\vspace{-8pt}
\begin{tabular}{lcccc}
\hline
\multicolumn{1}{c}{} &
  \begin{tabular}[c]{@{}c@{}}Requires \\ Replay \end{tabular} &
  \begin{tabular}[c]{@{}c@{}}Expands \\ Network\end{tabular} &
  \begin{tabular}[c]{@{}c@{}}Selects \\ Connections\end{tabular} &\\ \hline
GSS                       & \checkmark & $\times$ &  $\times$ \\
CoPE                     & \checkmark & $\times$ &   $\times$ \\
MIR                     & \checkmark& $\times$ &    $\times$ \\
CN-DPM                 & \checkmark & \checkmark &  $\times$ \\
Dynamic-OCM      & \checkmark & \checkmark &  $\times$ \\
SEDEM                  & \checkmark & \checkmark &   $\times$ \\
\textbf{SERENA (ours)}     & $\times$ & $\times$ & \checkmark \\ \hline
\end{tabular}%
\end{wraptable}

\paragraph{Background and Motivation.}
Various methods proposed in online data-incremental learning can be divided into two main branches: replay-based and architecture-based. Replay-based CL approaches store a subset of the previous examples to be able to adjust the features and decision boundaries for all classes. Architecture-based approaches, on the other hand, learn the features of each concept or session on an entirely new model. However, they (i) bring overhead costs in terms of both memory and computation, (ii) violate privacy concerns, and (iii) complicate data retrieval and training. 
This leads us to an essential question in online data-incremental learning:

\vspace{-2pt}
\begin{center}
\colorbox{gray!15}{ 
\parbox{0.97\textwidth}{
\centering
\vspace{0.5mm}
\emph{How can we enable continual learning without relying on complex procedures like initializing entirely new models or storing replay data while still promoting adaptation to new concepts and minimizing catastrophic forgetting?}
\vspace{0.5mm}
}
}
\end{center}

\vspace{-2pt}
To address this question, we benefit from the ability of deep neural networks to efficiently maintain multiple subnetworks or neural paths~\citep{ltr, topology}, allowing them to learn multiple tasks in a single model. This dynamic allocation of network connections enables models to preserve knowledge while adapting to new information, a crucial requirement in continual learning. 
Notably, this mechanism mirrors the brain’s selective activation strategy, where specific neuron regions respond to particular stimuli~\citep{brain}. Such selective responses facilitate efficient knowledge retrieval, prevent interference and aligns with \textit{‘self-regulated neurogenesis’}~\citep{neurogenesis} and Hebbian Theory’s principle~\citep{fire_wire} of \textit{‘Neurons that fire together, wire together,’} which suggests that frequently co-activated neurons strengthen their connections over time, reinforcing task-specific knowledge in certain neural paths or circuits~\citep{conceptcell}. 
By leveraging these principles, we aim to construct models that retain previously acquired representations while seamlessly integrating new knowledge into a single model, thereby bridging the gap between artificial and biological learning systems.

\begin{wraptable}[17]{l}{0.45\textwidth}
\captionsetup{font=small}
\label{tab:algo_train}
\renewcommand{\arraystretch}{1.2}
\small
\centering
\vspace{-12pt}
\begin{tabular}{p{0.9\linewidth}}
\\ \hline
\textbf{Algorithm 1: Pseudocode for SERENA} \\
\hline
\begin{algorithmic}
\STATE {\textbf{Input:}} Data stream set $\mathcal{S}$ with Batches of $\mathcal{B}$
\STATE {\hspace{2.8em}} Model {$\theta$}
\STATE {\hspace{2.8em}} Learning rate $\eta$
\STATE {\hspace{2.8em}} Model sparsity ratio $m$
\STATE {\hspace{2.8em}} Accuracy window $w$
\STATE {\hspace{2.8em}} Accuracy threshold $t$
\vspace{5pt}
\STATE{Eq. \ref{equ:erk} to create sparse expert $\theta_m \subset \theta$}
\vspace{0.5pt}
\FOR{$(x_i,y_i) \in \mathcal{B}$}
\STATE{Eq. \ref{equ:optim} to train $\theta_{m}$ with $\eta$}
\IF{accuracy of $\mathcal{B} \leq \mu_w \cdot t$}
\STATE{freeze $\theta_m$}
\STATE{Eq. \ref{equ:erk} to create new $\theta_m \subset \theta$}
\ENDIF
\ENDFOR
\vskip -0.2cm
\end{algorithmic}
\\ \hline
\end{tabular}
\end{wraptable}

\paragraph{Preliminaries.}
SERENA operates under an online data-incremental learning scenario which involves the continuous updating of a deep neural network model, where the data samples $(x_i, y_i)$ arrive gradually from a set of non-iid data streams $\mathcal{S} = \{ s_1, s_2, \dots, s_s \}$ with processing batches of size $B$. 
Data sample $i$ is constituted by input feature $x_i \in \mathcal{X}$ and corresponding signal $y_i$ in which the output space for classification is a discrete set of classes $\mathcal{Y} \leftarrow \mathcal{Y}_{i-1} \cup y_i$.
A deep network is optimized to map input instances to the classifier space in a flowing data stream, denoted as $f_{\theta}:\mathcal{X}\rightarrow \mathcal{Y}$, where $\theta$ represents the network parameters.
In online data-incremental learning, samples correspond to a series of tasks without any identifiers, in contrast to task- or class-incremental setup where explicit task cues are necessary.

\vspace{-12pt}
\begin{align}
\label{equ:erk}
1 - \frac{(n^{l-1}+n^{l}+w^{l}+h^{l})}{n^{l-1}  n^{l}  w^{l} h^{l}} 
\end{align}

\vspace{-5pt}
\paragraph{Concept Cell Allocation.}
SERENA uses zero-cost random unstructured pruning to the kernels of convolution layers and the connections of classifier head for each stream, using Erd\H{o}s-R\'{e}nyi Kernel (ERK)~\citep{rigl} as it enables better parameter efficiency, preserves model capacity in key regions, while suitable for longer data streams or larger number of tasks as demonstrated in prior work~\citep{cl_with_dst}.
Inspired by random graph theory, ERK allocates non-zero weights proportionally across layers based on the number of parameters, input/output channels and kernel dimensions, ensuring denser connectivity in the smaller layers. This allocation avoids over-pruning in the bottlenecks and has been shown to consistently outperform uniform sparsity. Formally, to obtain concept cell or neural path $\theta_m$ with overall sparsity level of binary mask $m$, it scales the sparsity of layers proportionally as given in Eq.~\ref{equ:erk}, where $n^l$ represents the number of neurons or feature maps at layer $l$, and $w^l$ and $h^l$ denote the width and height of the $l^{th}$ convolutional kernel, aiming to preserve the information flow.

\vspace{-3pt}
\paragraph{Drift Detection.}
SERENA incorporates a drift detection mechanism to continuously monitor the accuracy curve, enabling it to detect changes in the underlying concept being learned, similar to~\citep{ddm, adwin}. It tracks the accuracy of each batch of data $\mathcal{B}$ and compares it to the running accuracy over a defined window $w$. If the accuracy of the current batch $\mathcal{B}$ drops by more than a predetermined threshold $t$ relative to the mean of running accuracy, SERENA interprets this as a significant deviation or concept change. In response, it creates a new neural path $\theta_m$ within the same fixed backbone $\theta$ to adapt to the evolving concept.

\vspace{-3pt}
\paragraph{Training Flow.}
SERENA learns a different neural path or concept cell $\theta_m$ for each data stream $s_s$ by minimizing the function given in Eq.~\ref{equ:optim} in an online manner. 
Here, $CE$ denotes the Cross-Entropy Loss, $\theta_m$ denotes the randomly initialized concept cell, and the optimal $\theta_m^*$ is obtained by solving this optimization problem.
When a new stream $\mathcal{S}$ is identified by a drift detector, it freezes these connections of $\theta_m^*$ against forgetting and allocates a new concept cell $\theta_m$ to train on, which can share connections with other concept cells. This allows using the predefined main backbone $\theta$ more efficiently while enabling knowledge transfer between the concept cells. We share the details in Algorithm~1.

\vspace{-7pt}
\begin{align}
\label{equ:optim}
\theta_m^* &= \arg\min_{\theta_m} \sum_{i=1} [CE(f_{\theta_m}(x_{i}), y_{i})]
\end{align}

\vspace{-7pt}
\paragraph{Inference Flow.}
In evolving data streams, relying on a single selected path for classification can become unreliable due to distribution shifts, especially in the absence of explicit task or stream identifiers. Inspired by neurological processes, SERENA addresses this challenge through a novel ensemble classification strategy that mirrors how the brain tends to rely more heavily on the most recent experiences when making decisions, a phenomenon known as \textit{recency effect} \citep{recency_effect} or \textit{recency bias} \citep{recency_bias}.
Formally, each concept cell $\theta_m$ produces a logit vector $z_i = f_{\theta_m}(x) \in \mathbb{R}^\mathcal{|Y|}$. SERENA then computes the final logits $\tilde{z}$ as a weighted sum of these outputs, as shown in Eq. \ref{equ:inference}, in which a linearly increasing weighting scheme prioritizes more recent neural paths while still integrating past knowledge.
By dynamically adjusting the contributions of each concept cell, SERENA stabilizes decision boundaries and maintains robust predictions even without explicit task or stream labels. This neuro-inspired mechanism allows SERENA to be particularly effective in scenarios where pinpointing the correct neural path is inherently challenging, as its ensemble approach offers a reliable solution for managing evolving data distributions.

\vspace{-12pt}
\begin{align}
\label{equ:inference}
\tilde{z} = \sum_{i=1}^\mathcal{S} \frac{i}{\mathcal{S}}z_i, \quad \hat{y} = \arg \max_{y \in \mathcal{Y}} \tilde{z}_y
\end{align}



\section{Experiments}
\vspace{-5pt}
In this section, we describe the experimental setup and present experiments conducted on twelve different benchmarks to evaluate the incremental learning capabilities of SERENA in comparison with other state-of-the-art algorithms.
Additionally, we share a run-time \textit{vs.} accuracy analysis as well as a space projection of the samples to provide the robustness of the proposed method in more detail.

\subsection{Experimental Setup}
\paragraph{Datasets and Scenarios.}
In balanced scenarios, \textbf{Split-MNIST} and \textbf{Split-CIFAR10} have 5 streams, and each stream has 2 classes with 12000 samples for MNIST~\citep{mnist} and 10000 samples for CIFAR10~\citep{cifar} in training and 2000 samples in testing.\textbf{Split-CIFAR100} and \textbf{Split-MiniImageNet} divide CIFAR100~\citep{cifar} and MiniImageNet~\citep{miniimagenet} into 20 streams, each with 5 disjoint classes with 2500 samples for training and 500 samples for testing. \textbf{Split-TinyImageNet} divides TinyImageNet into 100 streams, each consist of 2 classes with 1000 samples for training and 100 samples for testing. Finally, \textbf{Split-ImageNet}~\citep{imagenet} has 20 streams, each with 50 classes with around 60000 samples for training and 2500 samples for testing.
In imbalanced scenarios, sample sizes exhibit a logarithmic increase or decrease pattern. \textbf{Ascending Split-MNIST} gradually rises from 2500 to 12000 samples per stream. \textbf{Ascending Split-CIFAR10} gradually increases from 2000 to 10000 samples per stream. \textbf{Ascending Split-CIFAR100} starts from 500 up to 2500 samples per stream. The descending counterparts; namely \textbf{Descending Split-MNIST}, \textbf{Descending Split-CIFAR10}, and \textbf{Descending Split-CIFAR100} mirror this trajectory in reverse.

\vspace{-2pt}
\paragraph{Compared Baselines.}
We compare with \textbf{finetune} as a naive approach; \textbf{reservoir}~\citep{reservoir}, \textbf{GSS}~\citep{gss}, \textbf{MIR}~\citep{mir}, \textbf{CoPE}~\citep{cope} as replay-based approaches; and \textbf{CN-DPM}~\citep{cndpm}, \textbf{Dynamic-OCM}~\citep{dynamic_ocm}, and \textbf{SEDEM}~\citep{sedem} as architecture expansion methods. Moreover, similar to \citep{cope, cndpm}, we compare with \textbf{iid-online} and \textbf{iid-offline} as upper reference points for performance by relaxing the non-iid property of CL to a traditional supervised learning setting where \textit{online} trains all classes at once for a single epoch and \textit{offline} for 50 epochs.
Note that TIL and CIL methods are not designed or suitable for this setup; thus, they are not included as direct baselines for comparison.

\vspace{-2pt}
\paragraph{Backbones.}
For Split-MNIST, we use a 2-layer MLP with 400 neurons in each layer before the classifier. We use standard a ResNet18~\citep{resnet} for the rest.
After each stream, we freeze used connections except batch normalization since they cannot be designated as stream-specific components.

\vspace{-2pt}
\paragraph{Evaluation Metrics.}
We measure model performance with common CL metrics which are Average Accuracy and Forgetting~\citep{aser, dvc}. 
Average Accuracy is computed by training the model on the data stream comprising all streams and subsequently evaluating the final model's performance using the test data associated with all streams. It is formally defined as: $\frac{1}{S} \sum_{j=1}^S a_{S,j}$ where $a_{i,j}$ is the accuracy on stream $j$ after the model is trained from stream 1 to $i$.
Average Forgetting quantifies the extent to which the model forgets previously learned streams after being trained on the final stream and can be defined as: $\frac{1}{S-1} \sum_{j=1}^{S-1} f_{S,j}$ where $f_{i,j} = \max_{k \in(1, 2, ..., i-1)} a_{k,j} - a_{i,j}$.

\vspace{-2pt}
\paragraph{Implementation Details.}
We train from scratch with a batch size of $10$, following~\citep{aser, ocm, onpro}. We set the window size to 10 and the accuracy drop threshold to $50\%$ for the drift detector. We use SGD with a learning rate of $1\times10^{-2}$ for MLP and $5\times10^{-4}$ for ResNet18 and weight decay of $5\times10^{-4}$ for both. When training new concept cells, we activate a fixed percentage of the weights, 20\% for MLP and 5\% for ResNet18. For the baselines, we use the same backbone and their default settings for a fair comparison. For online-joint, we train all classes at once for $1$ epoch with a batch size of 10 and for offline-joint, $50$ epochs with a batch size of 128.

\subsection{Results and Analysis}

\begin{figure*}[h]
\captionsetup{font=small}
  \centering
  \begin{subfigure}{0.328\textwidth}
    \caption{Split-MNIST (MLP)}
    \includegraphics[width=\textwidth]{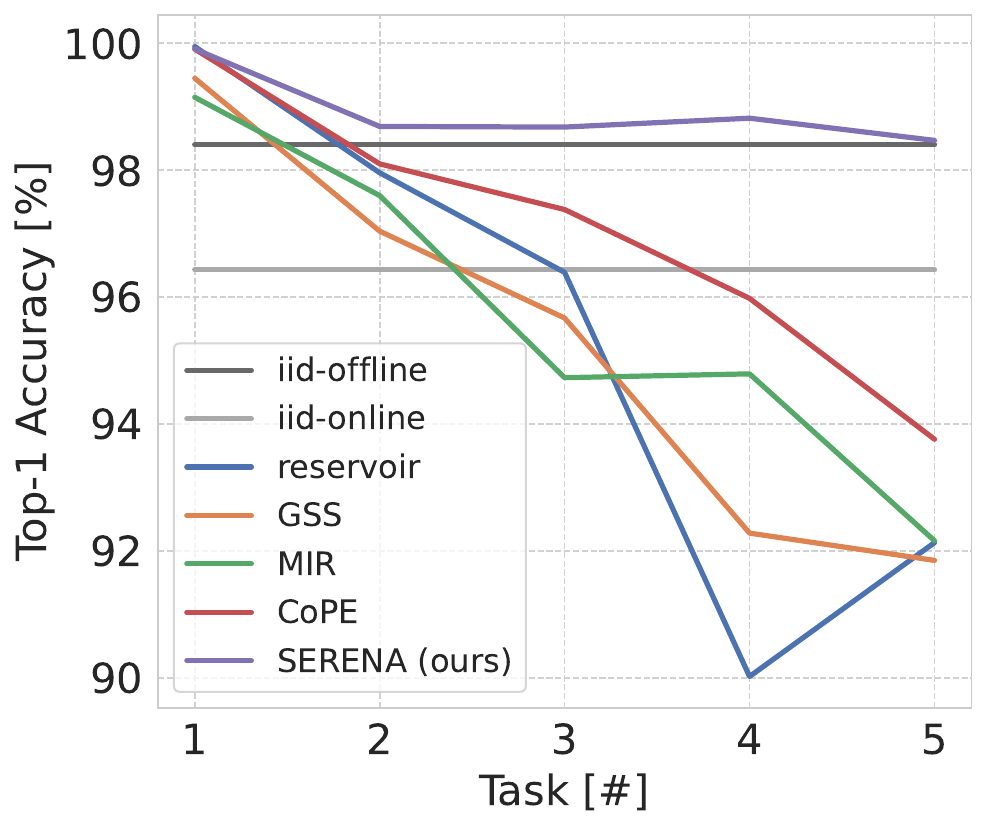}
  \end{subfigure}
  \begin{subfigure}{0.32\textwidth}
    \caption{Split-CIFAR10 (ResNet18)}
    \includegraphics[width=\textwidth]{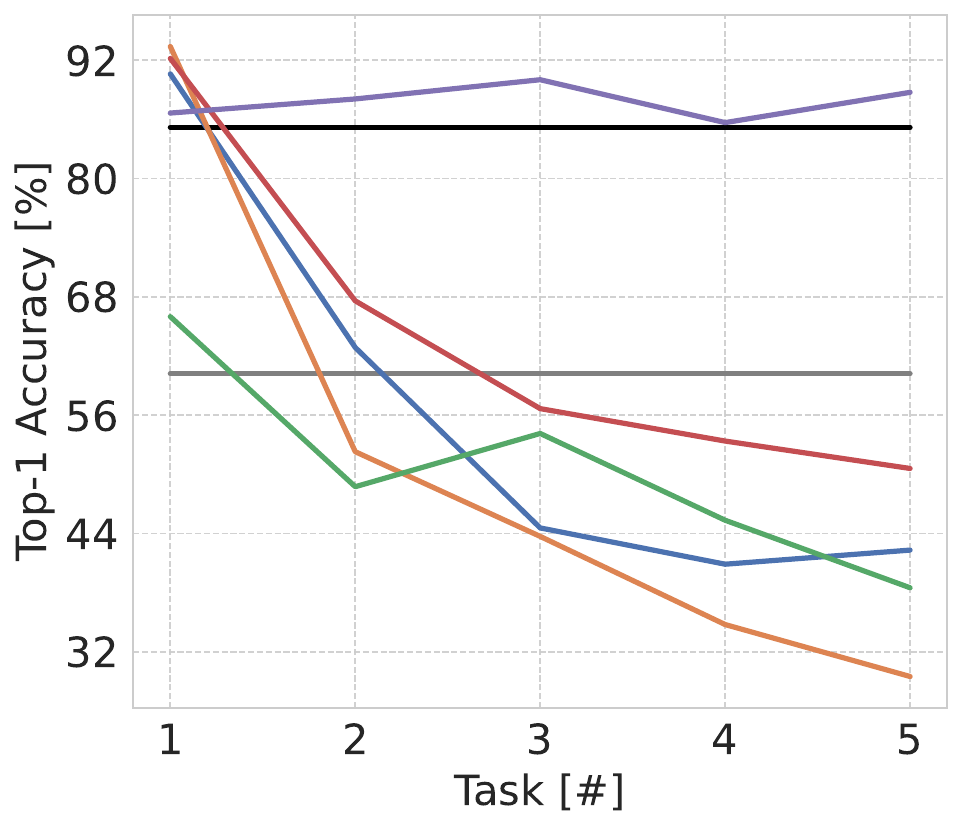}
  \end{subfigure}
  \begin{subfigure}{0.32\textwidth}
    \caption{Split-CIFAR100 (ResNet18)}
    \includegraphics[width=\textwidth]{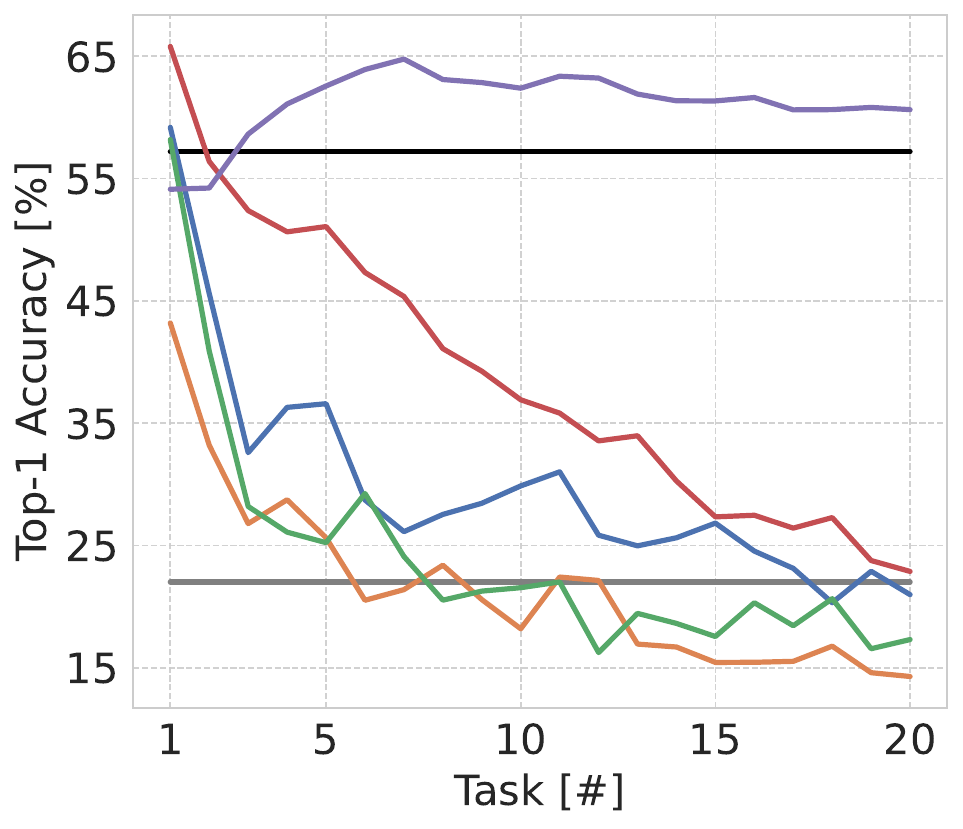}
  \end{subfigure}
  \vspace{-5pt}
\caption{\textbf{Average accuracy [\%]} after each stream on balanced benchmarks. SERENA outperforms all methods by a large margin, including iid-offline which is not a CL method but a standard supervised batch learning, representing an upper reference point.}
\label{fig:acc}
\vskip -0.4cm
\end{figure*}

\begin{figure*}[h]
\captionsetup{font=small}
  \centering
  \begin{subfigure}{0.31\textwidth}
    \caption{GSS}
    \includegraphics[width=\textwidth]{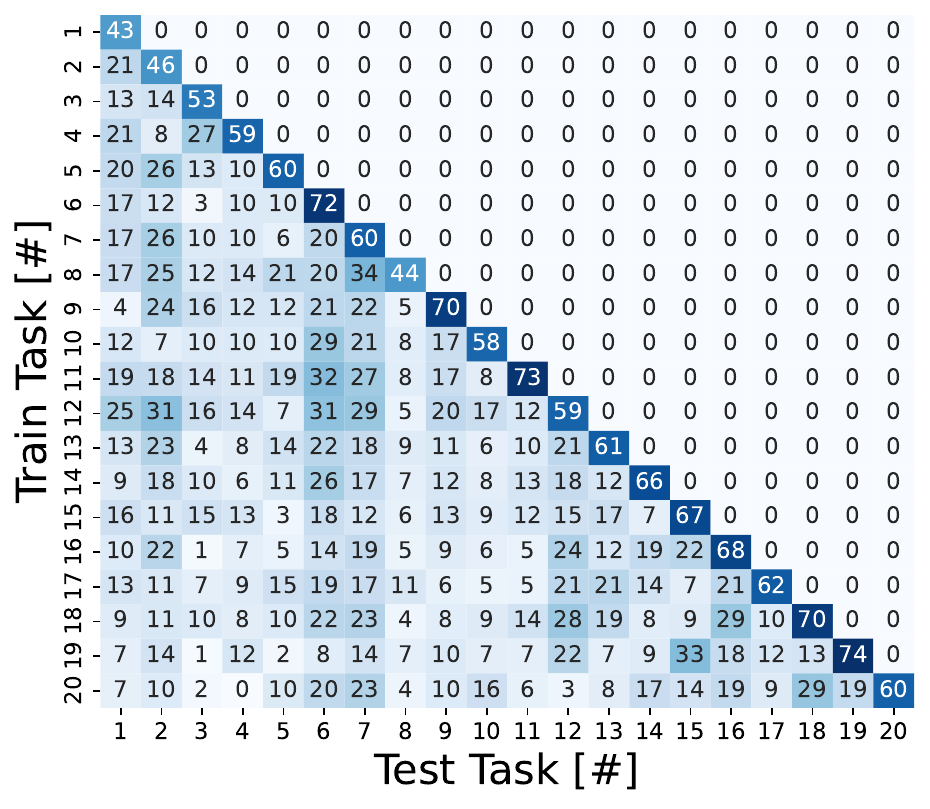}
  \end{subfigure}
  \begin{subfigure}{0.31\textwidth}
    \caption{CoPE}
    \includegraphics[width=\textwidth]{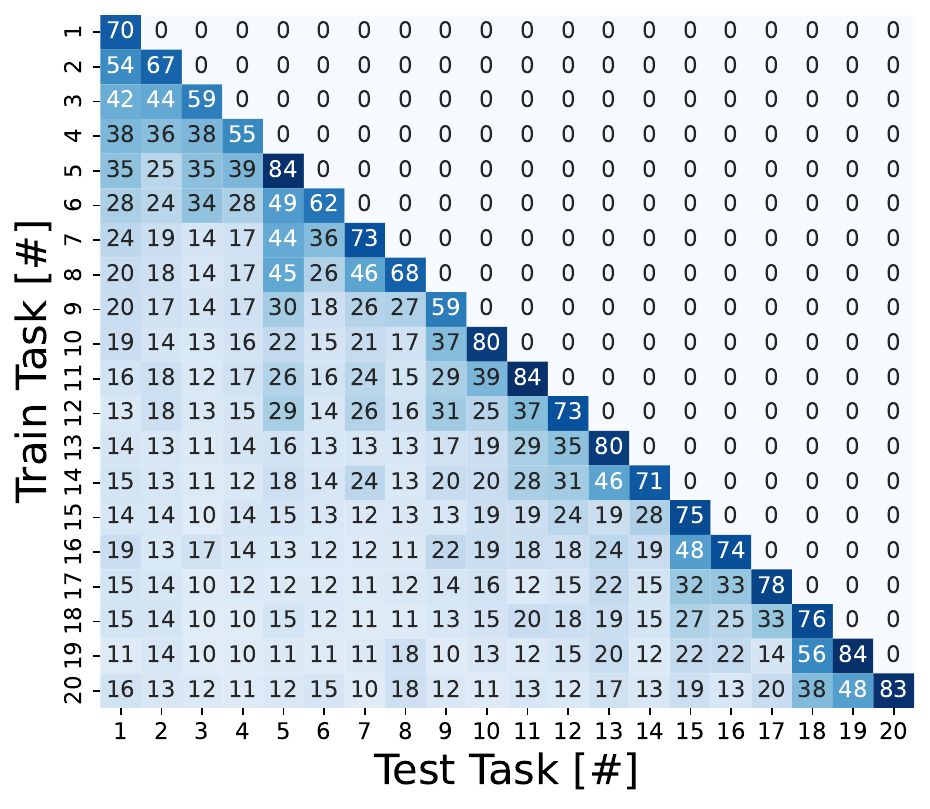}
  \end{subfigure}
  \begin{subfigure}{0.31\textwidth}
    \caption{SERENA}
    \includegraphics[width=\textwidth]{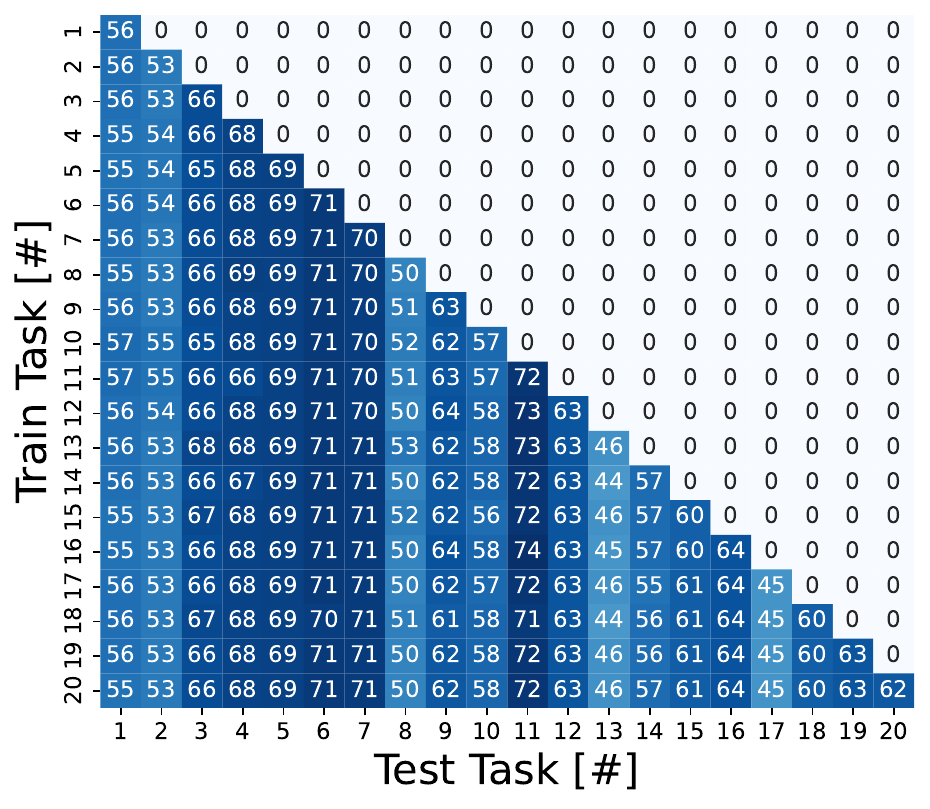}
  \end{subfigure}
\caption{\textbf{Accuracy [\%]} of each stream after the model has been trained on a new one for Split-CIFAR100. SERENA demonstrates its strength by maintaining consistently high accuracy and near-zero forgetting across the streams compared to existing approaches.}
\vskip -0.2cm
  \label{fig:heatmap}
\end{figure*}

\paragraph{Performance on balanced scenarios.}
In Table \ref{tab:accuracy}, the results demonstrate the superior performance of SERENA relative to state-of-the-art methods, highlighting its ability to learn continually. Furthermore, as illustrated in Figure \ref{fig:acc}, SERENA consistently achieves higher accuracy at every time step across all benchmarks, even in the absence of replay data (\textit{M=0}) and network expansion. 
A key factor behind SERENA’s success is its biologically inspired learning mechanism, which balances \textit{plasticity} and \textit{stability}. Existing approaches are generally good at plasticity by acquiring the knowledge of a given task while they suffer on the stability side by forgetting previously learned concepts, which can be observed in Table \ref{tab:forget} and Figure \ref{fig:heatmap}. 
Unlike existing state-of-the-art methods, which struggle to maintain past knowledge even with the existence of replay and architecture expansion, SERENA achieves this naturally, ensuring strong retention of prior concepts without compromising adaptation to new streams.
\begin{table*}[h]
\captionsetup{font=small}
\centering
\caption{\textbf{Average accuracy [\%]} (higher is better) of SERENA compared to existing methods on different benchmarks. The results of architecture expansion approaches are cited from~\cite{sedem} and denoted with †. Note that iid-offline and iid-online are not CL methods, and the best results are highlighted in bold.}
\label{tab:accuracy}
\fontsize{1.8}{2.4}\selectfont
\setlength{\arrayrulewidth}{0.05pt} 
\resizebox{1\textwidth}{!}{
\begin{tabular}{lrrrr}
\hline
Method &
  \multicolumn{1}{c}{\begin{tabular}[c]{@{}r@{}}Split-MNIST\\ \textit{(M=2k)}\end{tabular}} &
  \multicolumn{1}{c}{\begin{tabular}[c]{@{}r@{}}Split-CIFAR10\\ \textit{(M=1k)}\end{tabular}} &
  \multicolumn{1}{c}{\begin{tabular}[c]{@{}r@{}}Split-CIFAR100\\ \textit{(M=5k)}\end{tabular}}  &
  \multicolumn{1}{c}{\begin{tabular}[c]{@{}r@{}}Split-MiniImageNet\\ \textit{(M=5k)}\end{tabular}} \\ \hline
iid-offline        & 98.4 ± 0.1          & 85.2 ± 1.1          & 58.3 ± 0.6   & 54.7 ± 0.9    \\
iid-online         & 96.4 ± 0.2          & 59.9 ± 3.1          & 22.3 ± 3.2   & 19.2 ± 4.8   \\ \hline 
finetune             & 19.8 ± 0.1          & 17.6 ± 0.8          & 3.2 ± 0.3  & 3.7 ± 0.9  \\ 
reservoir            & 92.2 ± 0.7          & 42.7 ± 2.7          & 20.7 ± 2.8 & 7.0 ± 2.3  \\
GSS                  & 91.9 ± 1.8          & 29.2 ± 4.2          & 14.5 ± 2.9 & 6.9 ± 3.3  \\
CoPE                 & 93.8 ± 0.6          & 50.8 ± 2.1          & 22.8 ± 1.5 & 13.4 ± 1.1  \\
MIR                  & 91.8 ± 3.5          & 39.3 ± 5.2          & 17.1 ± 1.7 & 7.4 ± 2.3  \\ 
CN-DPM †             & 93.2 ± 0.1          & 45.2 ± 0.2          & 20.1 ± 0.1 & 27.9 ± 2.3  \\
Dynamic-OCM †        & 94.0 ± 0.2          & 49.2 ± 1.5          & 21.8 ± 0.7 & 26.6 ± 2.1  \\
SEDEM †              & 98.4 ± 0.2          & 55.3 ± 1.3          & 24.9 ± 1.2 & 29.6 ± 1.9 \\ \hline
\rowcolor{pink!20}
\textbf{SERENA} \textit{(M=0)} & \textbf{98.5 ± 0.1} & \textbf{88.7 ± 0.9} & \textbf{62.1 ± 1.7} & \textbf{48.7 ± 1.8} \\ \hline
\end{tabular}%
}  
\vskip -0.25cm
\end{table*}

Another remarkable advantage of SERENA is its ability to achieve performance levels comparable to iid-offline learning, making it particularly valuable for real-world applications where data is encountered sequentially. This capability bridges the gap between online continual learning and traditional offline training, demonstrating its potential for large-scale deployment. Notably, SERENA achieves this without initializing entirely new models or storing replay data, which is crucial for reducing computational costs, preserving privacy, and minimizing additional memory overhead.

\begin{wraptable}[12]{l}{0.46\textwidth}
\captionsetup{font=small}
\vskip -0.45cm
\caption{\textbf{Average accuracy [\%]} and \textbf{run time [hour]} on Split-ImageNet. SERENA reaches offline performance in an online data-incremental setup.}
\fontsize{8.5}{10.5}\selectfont
\setlength{\arrayrulewidth}{0.05pt}
\renewcommand{\arraystretch}{1.45}
\label{tab:imagenet}
\vskip -0.25cm
\begin{tabular}{llrr}
\hline
\multirow{1}{*}{Method} & Setup  & Split-ImageNet    & Run Time  \\   \hline
\multirow{2}{*}{iCaRL \textit{(M=20k)}}   & \textit{online}  & 2.11 & 6.6 hrs               \\
                                          & \textit{offline} & 14.96 & 64.2 hrs                \\
\multirow{2}{*}{FOSTER \textit{(M=20k)}}  & \textit{online}  & 0.94 & 11.4 hrs                 \\
                                          & \textit{offline} & 37.76 & 80.7 hrs                \\ \hline
\rowcolor{pink!20}
\textbf{SERENA} \textit{(M=0)}            & \textit{online}  & 36.22 & 4.1 hrs                \\ \hline
\end{tabular}
\end{wraptable}

\paragraph{Performance on a challenging scenario.} 
We evaluated SERENA also on the large-scale ImageNet dataset, as shown in Table \ref{tab:imagenet}, by dividing it into 20 sequential streams, a setup we refer to as Split-ImageNet. This benchmark serves as a realistic and challenging test for continual learning, as it requires models to learn from high-dimensional, diverse visual data without revisiting previous concepts. Due to the absence of direct baselines in existing codebases, we were unable to compare SERENA directly with them. Instead, we benchmark against iCaRL~\cite{icarl} and FOSTER~\cite{foster}, two widely recognized methods designed for offline class-incremental learning that provide implementations compatible with ImageNet. To ensure a fair comparison, we evaluated both methods using a replay buffer of 20 images per class (totaling 20,000 images) in two different scenarios: \textit{1-epoch online} and \textit{50-epoch offline}.
Our results reveal that iCaRL~\citep{icarl} and FOSTER~\citep{foster} struggle significantly in the online setup, which aligns with expectations, as both methods are inherently designed for offline continual learning, where multiple training epochs help mitigate forgetting. While SERENA is specifically designed for the online setting, it substantially outperforms a fundamental method in offline continual learning iCaRL~\citep{icarl} and achieves performance on par with the offline state-of-the-art method FOSTER~\citep{foster}, despite operating in a far more challenging single-pass learning scenario. These results underscore SERENA’s capacity to handle large-scale, real-world data streams efficiently without reliance on extensive memory buffers or multiple training iterations.

\begin{table*}[h]
\captionsetup{font=small}
\centering
\caption{\textbf{Average forgetting [\%]} (lower is better) of SERENA compared to existing methods on different benchmarks. The results of architecture expansion approaches are cited from~\cite{sedem} and denoted with †, and “–” indicates that the result was not reported. Note that iid-offline and iid-online return zero forgetting since they are supervised learners, not continual learners.}
\label{tab:forget}
\fontsize{1.8}{2.4}\selectfont
\setlength{\arrayrulewidth}{0.05pt} 
\resizebox{1\textwidth}{!}{
\begin{tabular}{lrrrr}
\hline
Method &
  \multicolumn{1}{c}{\begin{tabular}[c]{@{}r@{}}Split MNIST\\ \textit{(M=2k)}\end{tabular}} &
  \multicolumn{1}{c}{\begin{tabular}[c]{@{}r@{}}Split CIFAR10\\ \textit{(M=1k)}\end{tabular}} &
  \multicolumn{1}{c}{\begin{tabular}[c]{@{}r@{}}Split CIFAR100\\ \textit{(M=5k)}\end{tabular}} &
  \multicolumn{1}{c}{\begin{tabular}[c]{@{}r@{}}Split-MiniImageNet\\ \textit{(M=5k)}\end{tabular}} \\ \hline
iid-offline         & 0.0 ± 0.0           & 0.0 ± 0.0           & 0.0 ± 0.0            & 0.0 ± 0.0 \\
iid-online          & 0.0 ± 0.0           & 0.0 ± 0.0           & 0.0 ± 0.0            & 0.0 ± 0.0 \\ \hline 
finetune            & 79.5 ± 0.1          & 63.5 ± 2.2          & 58.6 ± 1.3           & 65.2 ± 2.1 \\
reservoir           & 6.4 ± 0.1           & 41.9 ± 2.1          & 44.2 ± 2.4           & 48.9 ± 2.2  \\
GSS                 & 6.9 ± 1.7           & 58.4 ± 4.9          & 45.3 ± 5.1           & 47.2 ± 4.7 \\
CoPE                & 4.3 ± 0.1           & 27.9 ± 2.8          & 36.6 ± 2.2           & 39.4 ± 2.4 \\
MIR                 & 6.8 ± 3.5           & 37.6 ± 9.8          & 39.9 ± 1.1           & 42.5 ± 1.5 \\
CN-DPM †            & –                   & –                   & –                    & – \\
Dynamic-OCM †       & –                   & –                   & –                    & – \\
SEDEM †             & –                   & –                   & –                    & – \\ \hline
\rowcolor{pink!20}
\textbf{SERENA} \textit{(M=0)} & \textbf{0.0 ± 0.0} & \textbf{0.7 ± 0.4} & \textbf{1.5 ± 1.1} & \textbf{1.7 ± 1.3} \\ \hline
\end{tabular}%
}
\vskip -0.2cm
\end{table*}

\begin{wraptable}[9]{l}{0.46\textwidth}
\captionsetup{font=small}
\caption{Used connections in ResNet18 indicate that a significant portion of connections remains inactive, suggesting ample capacity to accommodate additional data streams.}
\fontsize{9}{10}\selectfont
\setlength{\arrayrulewidth}{0.05pt}
\renewcommand{\arraystretch}{1.4}
\label{tab:network_saturation}
\vskip -0.1cm
\begin{tabular}{p{0.45\linewidth}p{0.45\linewidth}}
\hline
Dataset & Activated Parameters [\%] \\ \hline
Split-CIFAR10      &  \hspace{35pt}18 ± 2 \\
Split-CIFAR100     &  \hspace{35pt}51 ± 2 \\ 
Split-MiniImageNet &  \hspace{35pt}52 ± 3 \\ \hline
\end{tabular}
\end{wraptable}

\paragraph{Network Saturation.}
Random connection sharing among concept cells enables efficient network usage over a broader range of tasks. 
Table \ref{tab:network_saturation} illustrates that many connections stay inactive after processing all streams.
For example, Split-CIFAR10 consists of 5 streams, and 18\% of the network connections are utilized, despite each concept cell utilizing 5\% of the connections. Similarly, Split-CIFAR100 has 20 streams, and only 51\% of the connections are activated, showing that network usage does not increase linearly with the streams.

\begin{figure*}[h]
\captionsetup{font=small}
  \centering
  \begin{subfigure}{0.32\textwidth}
    \caption{Split-MNIST (MLP)}
    \vskip 0.1cm
    \includegraphics[width=\textwidth]{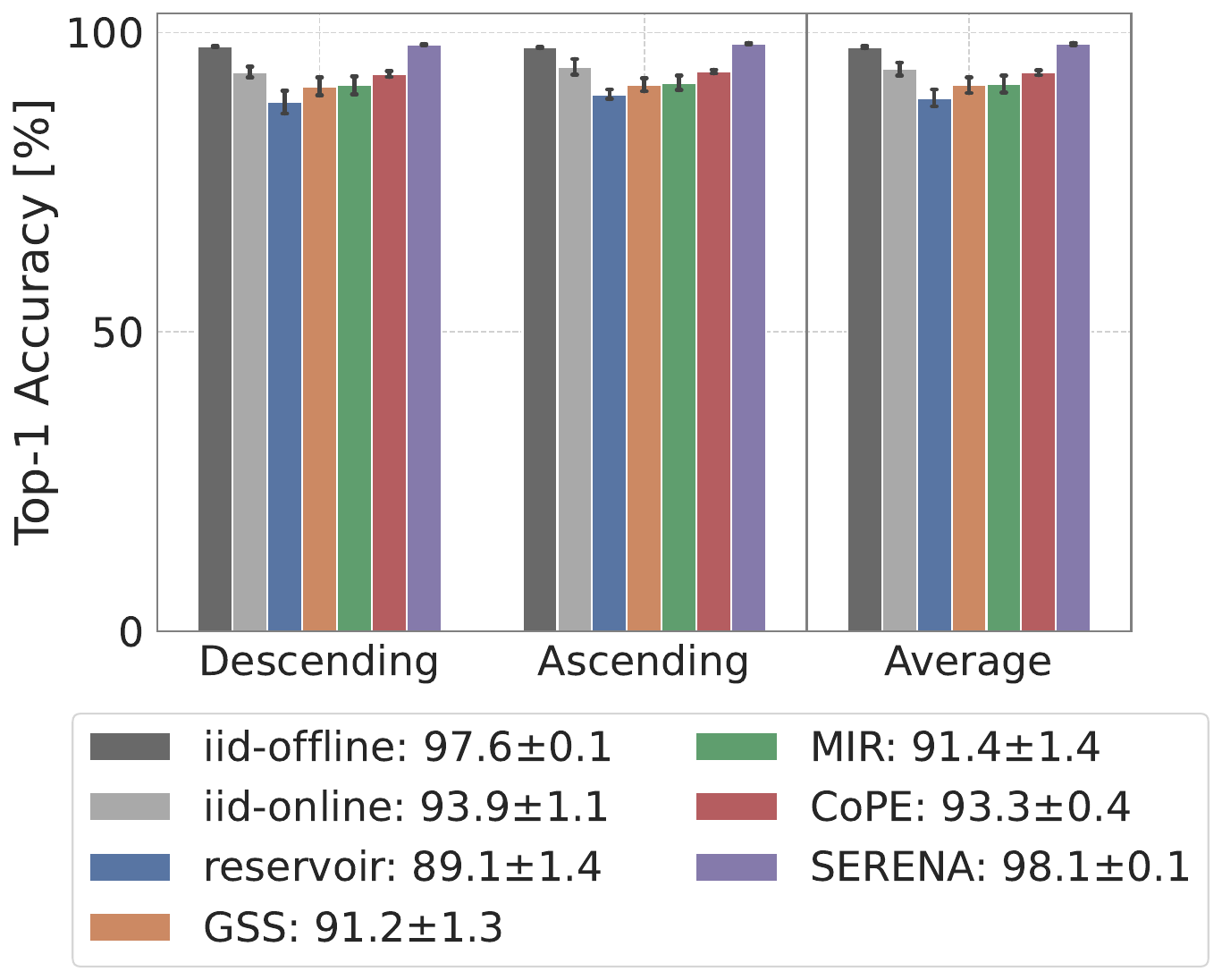}
    \label{fig:im_mnist}
  \end{subfigure}
  \begin{subfigure}{0.32\textwidth}
    \caption{Split-CIFAR10 (ResNet18)}
    \vskip 0.1cm
    \includegraphics[width=\textwidth]{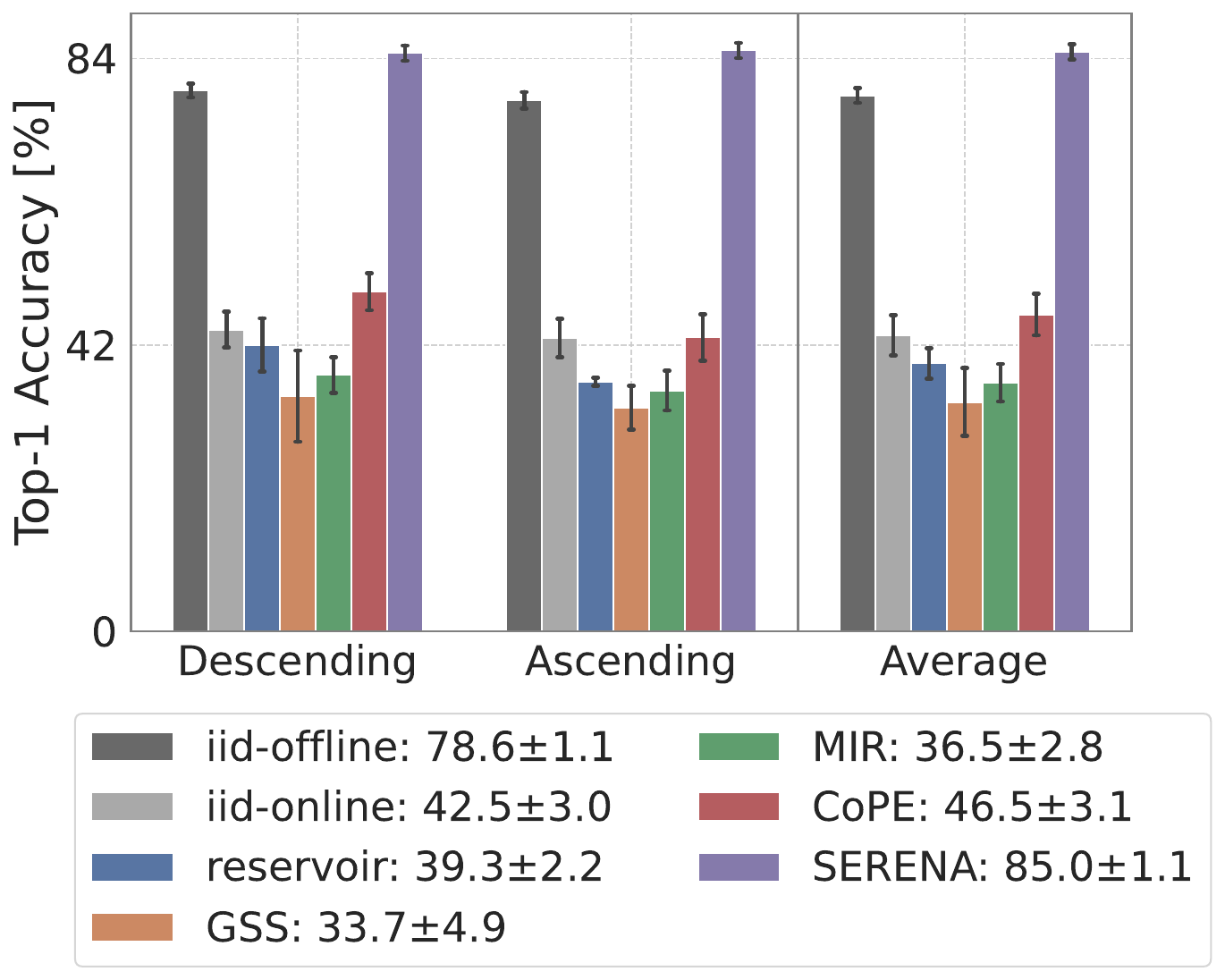}
    \label{fig:im_cifar10}
  \end{subfigure}
  \begin{subfigure}{0.32\textwidth}
    \caption{Split-CIFAR100 (ResNet18)}
    \vskip 0.1cm
    \includegraphics[width=\textwidth]{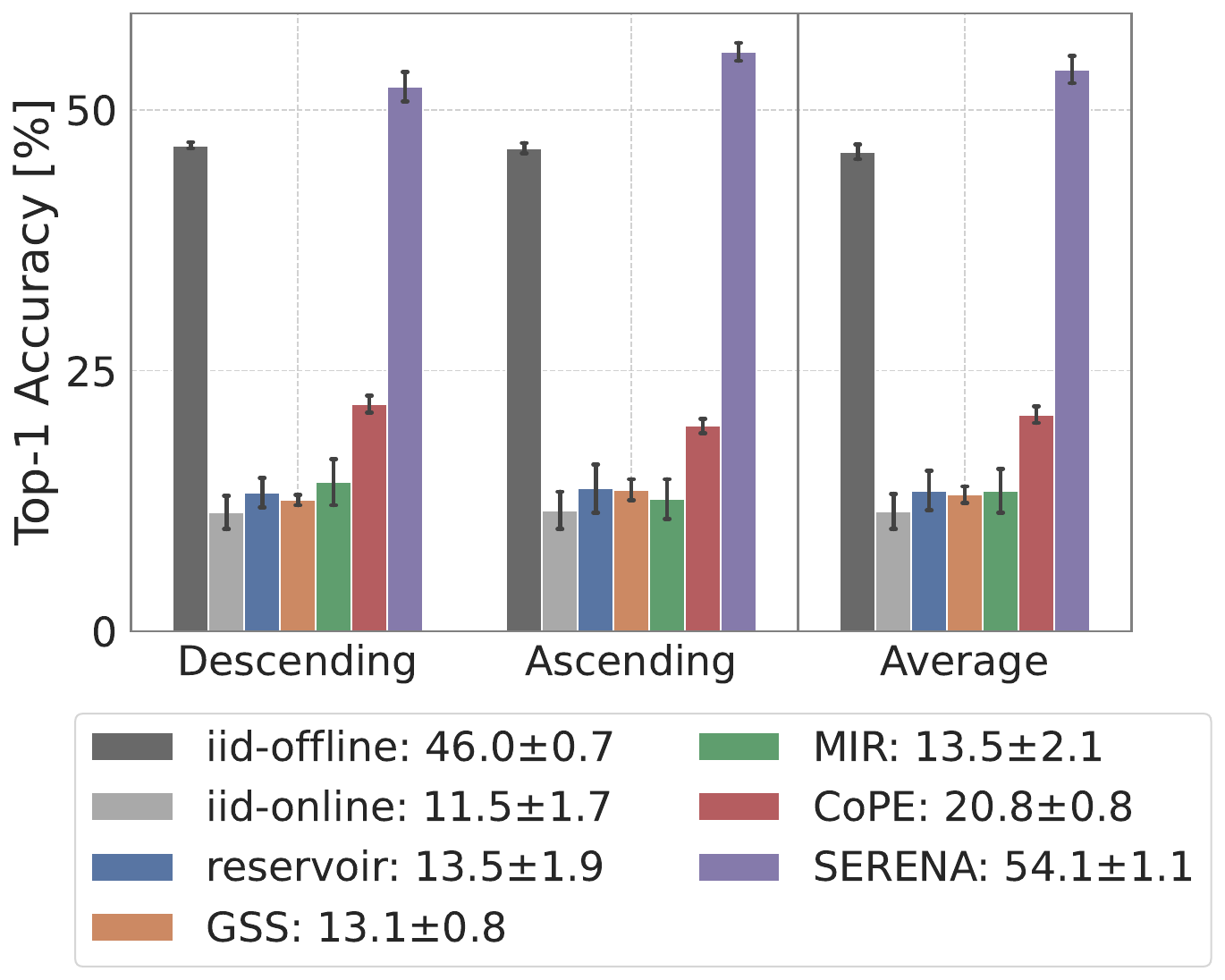}
    \label{fig:im_cifar100}
  \end{subfigure}
 \vspace{-5pt}
\caption{\textbf{Average accuracy [\%]} on different imbalanced benchmarks. Although all methods experience some level of performance degradation, SERENA improves existing ones by a large margin, including the upper reference point iid-offline, which is not a CL method but supervised batch learning.}
\label{fig:imbalance}
\vskip -0.2cm
\end{figure*}

\paragraph{Performance on imbalanced scenarios.}  
All methods experience a decline in average accuracy when exposed to the imbalanced data streams we introduce, with performance degradation becoming more pronounced as the degree of imbalance increases. However, SERENA demonstrates a significantly lower performance drop compared to other methods and consistently outperforms them, particularly in scenarios where the number of samples per stream is highly limited, as in Figure \ref{fig:im_cifar10} and Figure \ref{fig:im_cifar100}.  

This improvement can be attributed to SERENA’s novel mechanism of dynamically allocating new concept cells to incoming data streams. Unlike existing methods that struggle with class imbalance due to overemphasis on majority classes, SERENA effectively mitigates this bias by learning distinct representations on different neural paths for new data while preserving previously acquired knowledge. This ensures that rare or underrepresented classes receive sufficient attention, allowing for more balanced learning across different distributions.  

By maintaining equal focus across all classes and streams, SERENA not only reduces the detrimental effects of imbalance but also achieves performance levels comparable to or exceeding those of 50-epoch offline supervised batch learning. This robustness is particularly critical in real-world applications where data is often collected in an online fashion and naturally exhibits class imbalances, requiring models that can handle continuous long-term performance and imbalanced learning.

\begin{figure*}[h]
\centering
\begin{subfigure}{0.235\textwidth}
\captionsetup{font=footnotesize}
\caption{Split-MNIST}
\includegraphics[width=\textwidth]{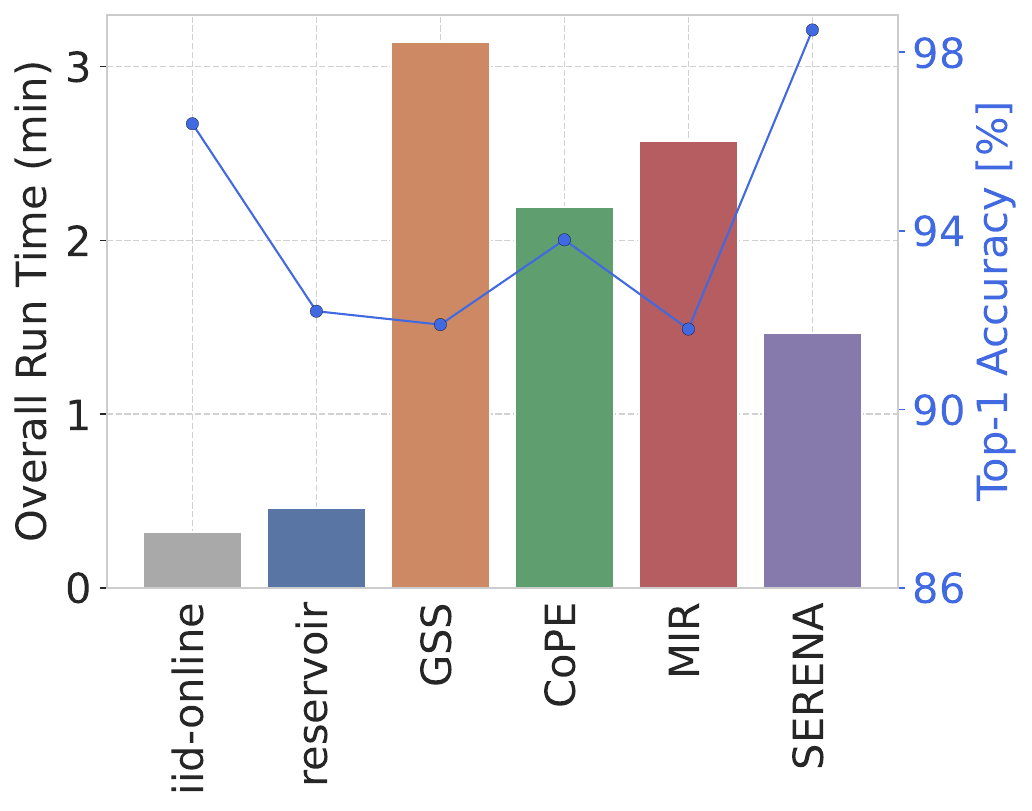}
\end{subfigure}
\hspace{2pt}
\begin{subfigure}{0.235\textwidth}
\captionsetup{font=footnotesize}
\caption{Split-CIFAR10}
\includegraphics[width=\textwidth]{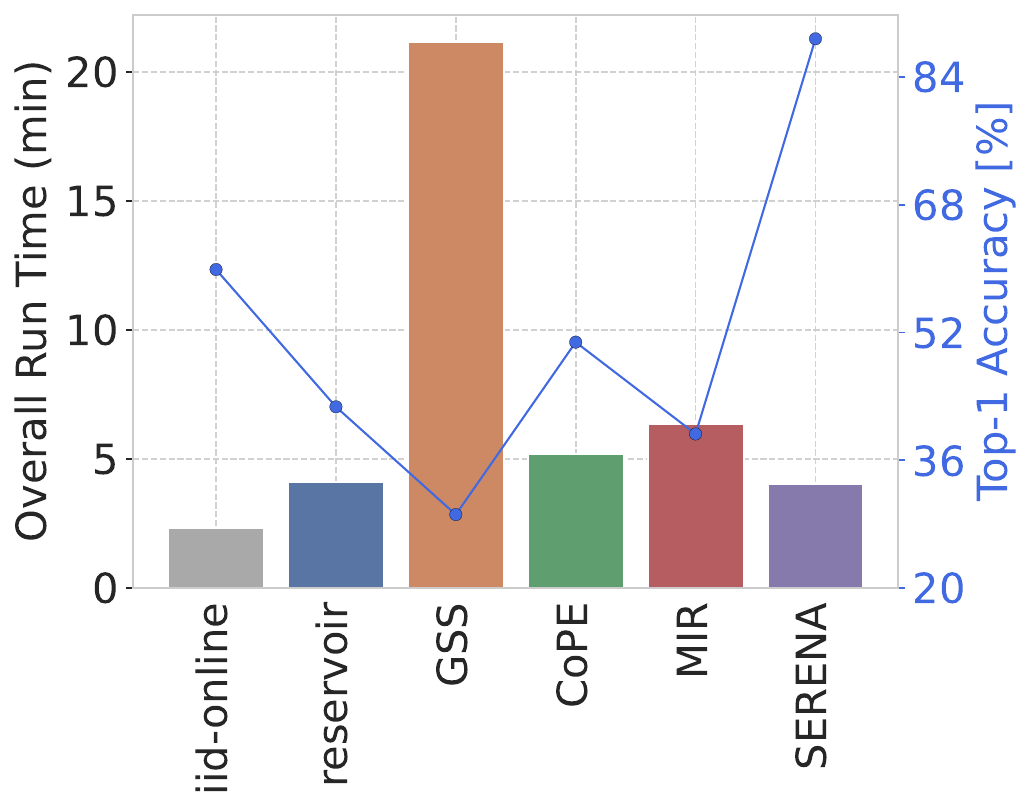}
\end{subfigure}
\hspace{2pt}
\begin{subfigure}{0.235\textwidth}
\captionsetup{font=footnotesize}
\caption{Split-CIFAR100}
\includegraphics[width=\textwidth]{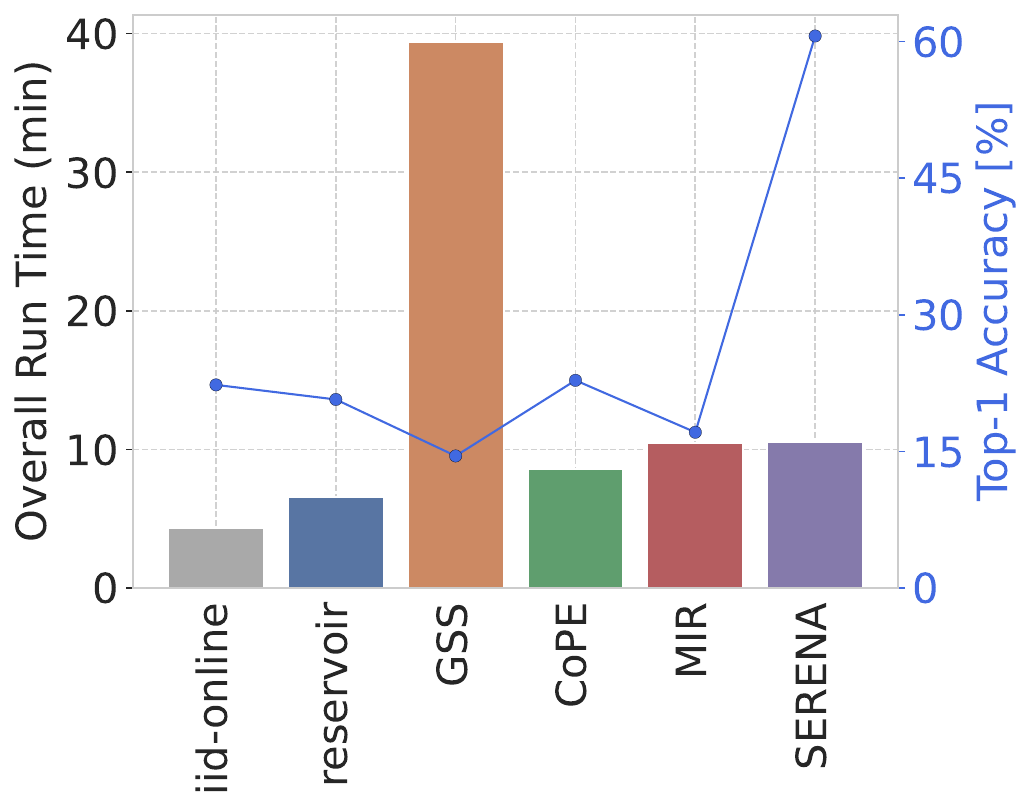}
\end{subfigure}
\hspace{2pt}
\begin{subfigure}{0.235\textwidth}
\captionsetup{font=footnotesize}
\caption{Split-MiniImageNet}
\includegraphics[width=\textwidth]{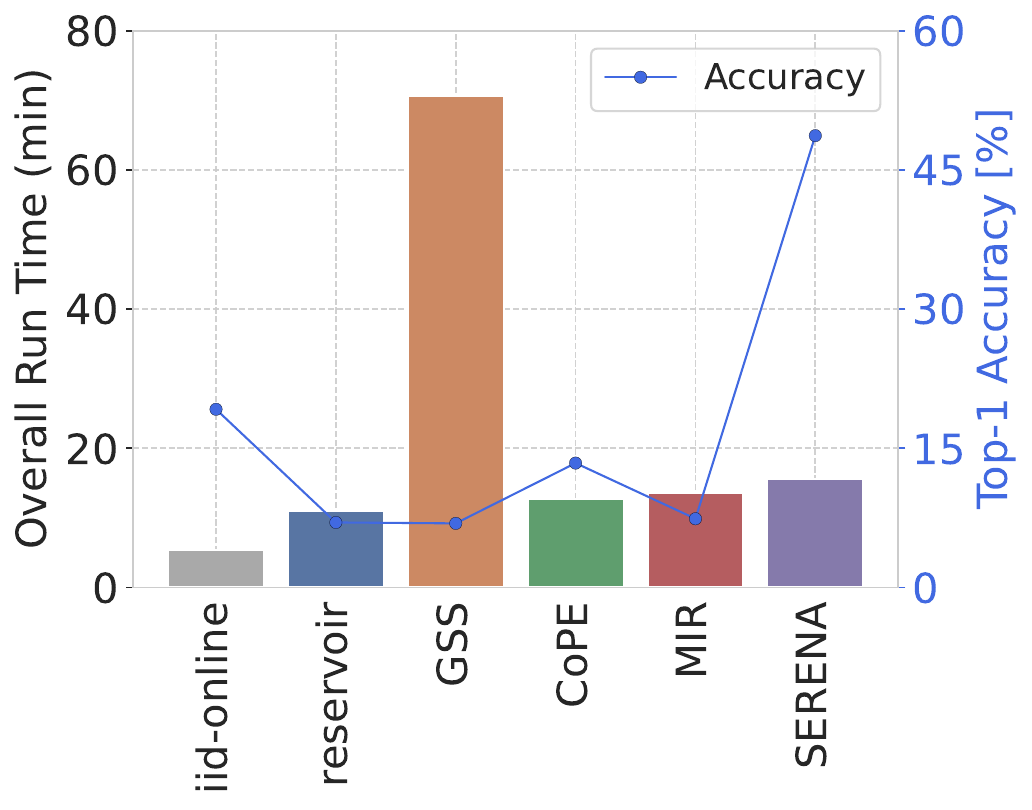}
\end{subfigure}
\captionsetup{font=small}
\caption{Overall runtime \textit{vs.} average accuracy for each method. SERENA maintains a comparable runtime to existing methods while achieving significantly higher accuracy, demonstrating its efficiency in online data-incremental learning scenarios.}
\label{fig:runtime}
\end{figure*}

\paragraph{Run Time.}  
SERENA maintains an on-par overall runtime to state-of-the-art methods while achieving significantly higher accuracy by utilizing only a small, randomly selected subnetwork for each stream. This selective activation reduces computational overhead during training and enhances efficiency in handling sequential learning tasks. For instance, on a single A100 GPU, SERENA completes Split-CIFAR10, Split-CIFAR100, and Split-MiniImageNet in approximately 5 minutes, 11 minutes, and 16 minutes respectively, matching the runtime of existing methods while nearly doubling the achieved accuracy, as illustrated in Figure \ref{fig:runtime}. This efficiency is particularly crucial in online learning environments where rapid adaptation to new data is essential. By optimizing resource utilization without sacrificing performance, SERENA presents a practical solution for real-time continual learning scenarios, making it well-suited for large-scale applications with strict computational constraints.

\paragraph{Ablation Study.}
We present the results in Table \ref{tab:ablate_hpo}, which provides a comprehensive analysis of the performance of our models under different configurations, specifically varying in terms of sparsity levels, window sizes, drift detection thresholds, and learning rates, all evaluated on the Split-CIFAR10 benchmark.
To better understand the sensitivity and robustness of our approach, we experimented with sparsity levels of $0.80$, $0.95$, and $0.99$, each tested across multiple window sizes of $5$, $10$, and $20$, and paired with drift detection thresholds of $0.1$, $0.5$, and $0.9$. Additionally, we varied the learning rate across a range of values, including $0.0005$, $0.001$, $0.005$, and $0.01$, to observe the interaction between learning dynamics and structural compression.

From these experiments, we observe that extreme sparsity levels, particularly $0.99$, or overly aggressive learning rates such as $0.01$ frequently lead to a noticeable drop in final model accuracy. This suggests that excessive pruning or overly rapid updates result in unstable learning dynamics and hinder the model convergence to effectively capture meaningful patterns from the input data. On the other hand, a moderate sparsity with smaller updates tends to strike a favorable balance between model compactness and task performance.

Among all tested configurations, the best performance, reaching an accuracy of 89.7\%, was achieved using a sparsity level of $0.95$, a window size of $10$, a drift threshold of $0.5$, and a conservative learning rate of $0.0005$. We believe this particular combination provides a sweet spot where the model can adapt adequately to distributional shifts between tasks without overreacting to noise or losing generalization.

\begin{table}[h]
\captionsetup{font=small}
\caption{Ablation of different network sparsities, learning rates, window sizes, and threshold values on Split-CIFAR10 in terms of \textbf{average accuracy [\%]}. Best result is highlighted in bold.}
\label{tab:ablate_hpo}
\fontsize{14}{18}\selectfont
\setlength{\arrayrulewidth}{0.05pt} 
\resizebox{\textwidth}{!}{%
\begin{tabular}{ccccccccccc}
\hline
\multicolumn{2}{c}{Sparsity Level} & \multicolumn{3}{c}{0.80}             & \multicolumn{3}{c}{0.95}             & \multicolumn{3}{c}{0.99}             \\ \hline
Window Size    & Drift Threshold   & lr = 0.0005 & lr = 0.001 & lr = 0.01 & lr = 0.0005 & lr = 0.001 & lr = 0.01 & lr = 0.0005 & lr = 0.001 & lr = 0.01 \\ \hline
\multirow{3}{*}{5}  & 0.1 & 89.0 & 89.3 & 41.0 & 89.5 & 88.3 & 40.9 & 88.8 & 89.3 & 41.1 \\
                    & 0.5 & 89.5  & 88.8 & 41.1  & 89.5 & 88.5 & 41.5 & 88.9 & 89.1 & 40.9 \\
                    & 0.9 & 89.1 & 88.7  & 41.1  & 87.8  & 88.3 & 41.3 & 89.5 & 88.7  & 40.6 \\ \hline
\multirow{3}{*}{10} & 0.1 & 89.3 & 88.9  & 41.1 & 89.5 & 88.3   & 40.9 & 89.5 & 89.2& 40.8\\
                    & 0.5 & 89.4 & 89.4  & 40.1  & \textbf{89.7} & 89.0 & 41.1 & 89.2 & 88.7 & 41.1\\
                    & 0.9 & 88.8 & 88.8 & 41.0  & 89.6 & 87.6 & 39.9 & 89.5 & 89.1 & 41.2\\ \hline
\multirow{3}{*}{20} & 0.1 & 89.2 & 89.2  & 41.2 & 89.5  & 88.5 & 41.0 & 89.3 & 88.9 & 41.1\\
                    & 0.5 & 89.2  & 89.2  & 41.3  & 89.4 & 88.3 & 40.5  & 89.2 & 89.1 & 40.9\\
                    & 0.9 & 80.3  & 89.1 & 40.9  & 80.0  & 88.6 &  41.1 & 89.4 & 89.2 & 40.9\\ \hline
\end{tabular}%
}
\end{table}

\section{Discussion}
In this section, we first investigate how SERENA achieves performance comparable to offline learning by analyzing the representation space of the neural network. Furthermore, we examine two critical discussion points for continual learning: never-ending learning and class-revisiting scenarios. These settings pose unique challenges for fixed-capacity replay-free systems and demand novel strategies for efficient knowledge acquirement, retention, and reuse. Although not the central focus of our current experimental evaluation, we discuss how SERENA’s working principles offer a strong foundation for addressing these complex scenarios.

\begin{figure}[h]
\captionsetup{font=small}
  \centering
  \begin{subfigure}{0.28\columnwidth}
    \caption{iid-online}
    \vskip 0.1cm
    \includegraphics[width=\textwidth]{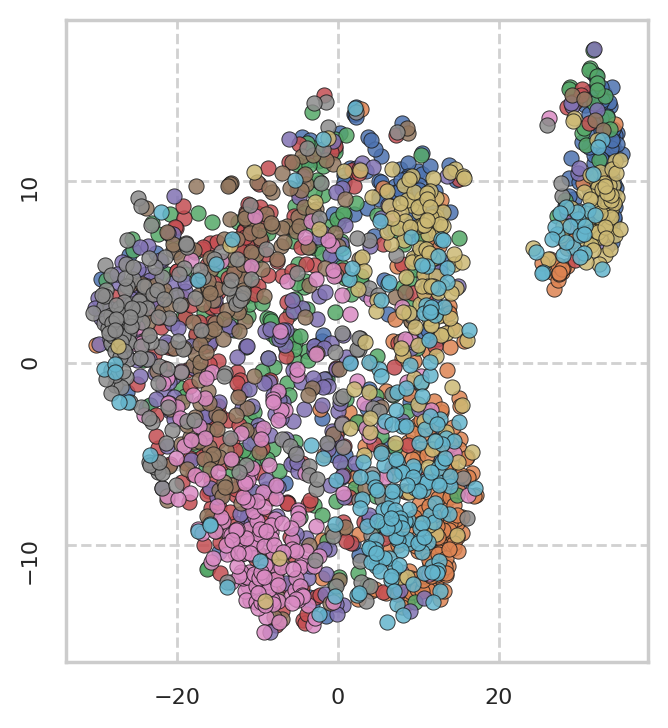}
  \end{subfigure}
  \hspace{20pt}
  \begin{subfigure}{0.28\columnwidth}
    \caption{iid-offline}
    \vskip 0.1cm
    \includegraphics[width=\textwidth]{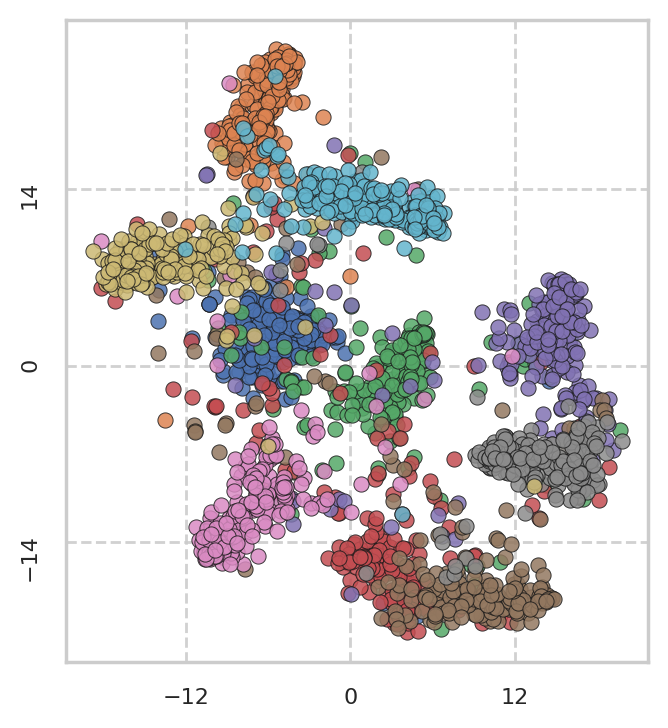}
  \end{subfigure}
  \hspace{20pt}
  \begin{subfigure}{0.28\columnwidth}
    \caption{SERENA (ours)}
    \vskip 0.1cm
    \includegraphics[width=\textwidth]{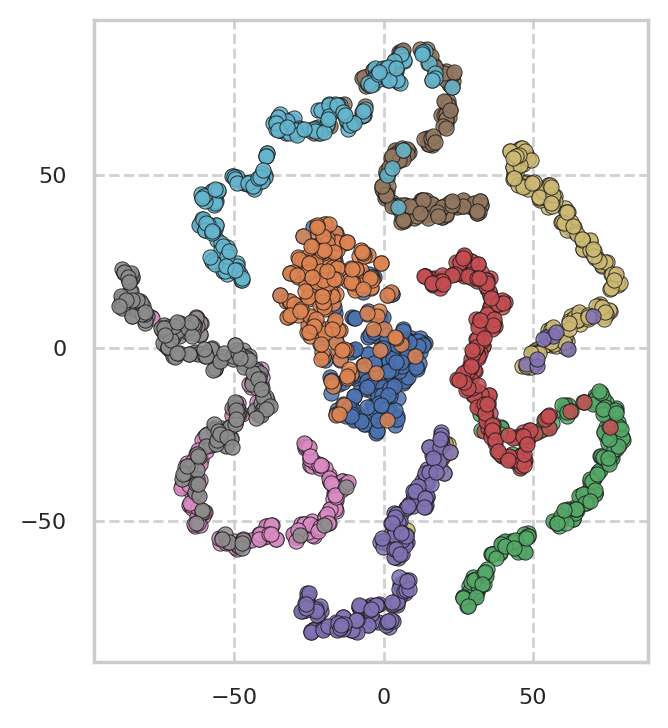}
  \end{subfigure}
\caption{By learning specialized neural paths or concept cells~\cite{conceptcell} in a single pass, SERENA effectively separates distinct data streams and classes--as indicated by different colors--with minimal overlap.}
\label{fig:tsne}
\vskip -0.2cm
\end{figure}

\paragraph{Space Projection.}  
To better understand why SERENA surpasses the upper reference point iid-offline, which uses 50-epoch supervised batch learning, we analyze the t-SNE projection of feature representations, \textit{after training on all streams}. This analysis provides insight into how well different ways of training separate the streams and the classes within the same streams. 
As shown in Figure \ref{fig:tsne}, the iid-online approach struggles to maintain distinct class boundaries, resulting in a near-random spread of samples, indicating poor feature separation. While the iid-offline approach improves upon this, showing clearer class differentiation, overlaps and moderate spread persist, increasing the risk of misclassification.  

In contrast, SERENA exhibits well-defined class groups, achieving superior separation of each stream. Any observed overlap is negligible and typically occurs only between semantically related classes within the same stream. The curved structures in the t-SNE plot indicate that each network path or concept cell has learned a distinct manifold on top of the initial stream, which forms a simpler and more compact cluster (orange-blue pair). This remarkable ability highlights SERENA’s effectiveness in structuring learned representations, making it highly resilient to forgetting. By dynamically leveraging distinct concept cells for each detected stream, SERENA ensures strong feature discrimination, facilitating continual learning while achieving the performance level of the offline supervised batch learning.

\begin{wraptable}[12]{l}{0.45\textwidth}
\captionsetup{font=small}
\caption{\textbf{Average accuracy [\%]} on Split-TinyImageNet, which simulates a never-ending scenario with 100 streams.}
\vspace{-5pt}
\label{tab:never_ending}
\fontsize{8.5}{8.9}\selectfont
\setlength{\arrayrulewidth}{0.05pt}
\renewcommand{\arraystretch}{1.4}
\begin{tabular}{lrr}
\hline
 & \begin{tabular}[c]{@{}r@{}}Split-TinyImageNet\\ (M=4k)\end{tabular} & \begin{tabular}[c]{@{}r@{}}Used Parameters\\ (from 11M)\end{tabular} \\ \hline
finetune     & 1.7  & 11M (100\%) \\
reservoir    & 2.0  & 11M (100\%) \\
GSS          & 2.4  & 11M (100\%) \\
CoPE         & 3.2  & 11M (100\%) \\
MIR          & 2.9  & 11M (100\%) \\ \hline
\rowcolor{pink!20}
SERENA (M=0) & \textbf{56.9} & 9.8M (89\%) \\ \hline
\end{tabular}%
\end{wraptable}

\paragraph{Never-Ending Scenario.}
Network saturation and reduced plasticity are inherent limitations of many fixed-resource replay-free continual learning methods, including SERENA.
Nevertheless, findings suggest that our approach offers a promising direction for such scenarios. Its design allows for efficient parameter utilization by reusing existing connections across streams rather than allocating entirely new parameters for each stream and this significantly delays the saturation without compromising performance. Our empirical analysis in Table~\ref{tab:network_saturation} substantiates this claim by demonstrating a sublinear growth in parameter usage by overlap across different neural paths. To emulate a never-ending scenario, we designed an experimental setup involving 100 sequential streams using TinyImageNet~\citep{tinyimagenet} in Table~\ref{tab:never_ending}. Notably, SERENA surpasses existing approaches both in performance and efficiency. While infinite stream lengths will inevitably exhaust any fixed model capacity, SERENA’s ability to sustain learning across long task sequences with minimal redundancy represents a significant step toward scalable, never-ending scenarios.

\begin{wrapfigure}[13]{l}{0.45\textwidth}
  \vspace{-17pt}
  \includegraphics[width=0.46\textwidth]{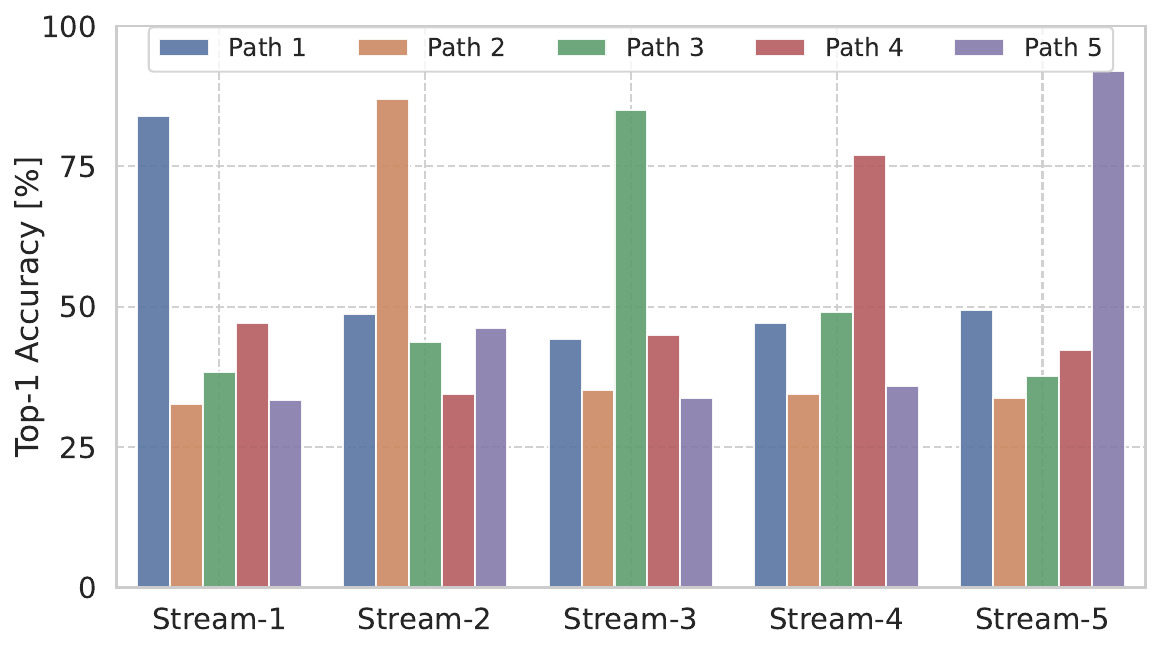}
  \captionsetup{font=small}
  \vspace{-20pt}
  \caption{Neural path performances on each stream, after completing all learning sessions on Split-CIFAR10.}
  \label{fig:revisit}
\end{wrapfigure}

\paragraph{Class-Revisiting Scenario.}
We conduct extensive evaluations of SERENA on widely recognized online continual learning benchmarks to ensure fair comparisons. While those benchmarks do not explicitly consider repetitive stream scenarios, we acknowledge that such settings represent a compelling and realistic extension of the continual learning. Despite their practical relevance, task revisiting scenarios are not yet standardized benchmarks in the field and thus remain mostly underexplored. Therefore, it is an important challenge that reflects a broader and largely unexplored area in continual learning. For example, replay-based methods store samples from all the tasks they encounter without discerning whether a task has truly changed or repeated. Similarly, architecture expansion methods typically allocate new parameters for each incoming task without verifying task novelty, resulting in unnecessary model growth.
Although SERENA does not currently implement a dedicated mechanism for task recurrence, its architectural design offers natural extensibility to such cases. Specifically, SERENA’s integration of concept drift detection and stream-specific neural paths enables the evaluation of existing paths upon encountering a new input stream. If a previously established path yields high predictive accuracy, as proved in Figure~\ref{fig:revisit}, it could be reused and interpreted as a sign of task recurrence rather than allocating additional resources. This potential for reuse not only mitigates unnecessary parameter growth but also highlights SERENA’s flexibility and efficiency, making it a strong candidate for future research in class-revisiting continual learning.

\vspace{-2pt}
\section{Conclusion}  
\vspace{-2pt}
Existing online data-incremental learning approaches rely heavily on complex mechanisms, such as storing replay data or expanding the model with entirely new network architectures, both of which introduce significant overhead costs and potential privacy concerns. Inspired by self-regulated neurogenesis~\citep{neurogenesis, neurogenesis1} and concept cells~\citep{conceptcell} observed in the human brain, we introduce SERENA for online data-incremental learning.
It dynamically adapts to concept drift by continuously fine-tuning the model and allocating neural connections or pathways for new concepts without storing past data or expanding the initial architecture. This enables SERENA to learn efficiently in a streaming setting while maintaining stability and minimizing interference. Additionally, we introduce new continual learning scenarios designed to better reflect real-world data dynamics where the sample sizes gradually increase or decrease across learning sessions.
Our extensive experiments demonstrate that SERENA not only surpasses state-of-the-art methods but also surpasses offline supervised batch learning performance. This underscores its effectiveness in real-world applications where storage constraints, computational efficiency, and privacy considerations are critical. By eliminating the need for replay data and avoiding unnecessary architectural expansion, SERENA provides a scalable and biologically inspired solution for continual learning.  

\textbf{Limitations and Future Work.}  
While SERENA efficiently utilizes network capacity, prolonged learning across an increasing number of streams may lead to model saturation, potentially limiting its ability to accommodate new tasks. Future work could explore adaptive mechanisms for optimizing network capacity and other continual learning scenarios, such as blurry and few-shot setups.

\section*{Broader Impact}
This paper presents work whose goal is to advance the field of Machine Learning, especially on the subject of Online Data-Incremental Learning. Besides the advancements in the field, it eliminates the need to store data or expand models, thereby diminishing privacy, memory, computation, and scalability concerns.

\section*{Acknowledgements}
This work is supported by SYNERGIES, a project funded by the EU Horizon programme under GA No. 101146542; Dutch national e-infrastructure with the support of SURF Cooperative using GA no. EINF-10242; and Turkish MoNE scholarship.


\bibliography{collas2025_conference}
\bibliographystyle{collas2025_conference}



\end{document}